\documentclass[11pt]{article}
\usepackage[final]{acl}
\usepackage{times}
\usepackage{latexsym}
\usepackage{amsmath}
\usepackage[T1]{fontenc}
\usepackage[utf8]{inputenc}
\usepackage{microtype}
\usepackage{inconsolata}
\usepackage{graphicx}
\usepackage{dirtree}
\usepackage{multirow}
\usepackage{multicol}
\usepackage{booktabs}
\usepackage{tabularx}
\usepackage{comment}
\usepackage{pgfplots}
\usepackage{acl}
\pgfplotsset{compat=1.18}
\usepackage{float}
\usepackage{xcolor}
\newcommand{\red}[1]{\textcolor{red}{#1}}

\definecolor{blue1}{RGB}{222, 235, 247}
\definecolor{blue2}{RGB}{198, 219, 239}
\definecolor{blue3}{RGB}{158, 202, 225}
\definecolor{blue4}{RGB}{107, 174, 214}
\definecolor{blue5}{RGB}{66, 146, 198}
\definecolor{blue6}{RGB}{33, 113, 181}
\definecolor{blue7}{RGB}{8, 69, 148}
\usepackage[utf8]{inputenc}
\usepackage[T1]{fontenc}
\usepackage{subcaption}

\title{OCRTurk: A Comprehensive OCR Benchmark for Turkish}

\author{
    Deniz Yılmaz\textsuperscript{1}, Evren Ayberk Munis\textsuperscript{2}, Çağrı Toraman\textsuperscript{1}, \\
    \textbf{Süha Kağan Köse}\textsuperscript{3}, \textbf{Burak Aktaş}\textsuperscript{3}, \textbf{Mehmet Can Baytekin}\textsuperscript{3},
    \textbf{Bilge Kaan Görür}\textsuperscript{3} \\
    \textsuperscript{1}Middle East Technical University, Computer Engineering Department, Turkey \\
    \textsuperscript{2}Politecnico di Torino, Italy \\
    \textsuperscript{3}Roketsan Inc., Artificial Intelligence Technologies Unit, Turkey \\
    \texttt{deniz.yilmaz\_12@metu.edu.tr}, \texttt{evrenayberk.munis@studenti.polito.it} \\
    \texttt{ctoraman@metu.edu.tr}, 
    \texttt{kagan.kose@roketsan.com.tr},
    \texttt{burak.aktas@roketsan.com.tr} \\
    \texttt{can.baytekin@roketsan.com.tr},
    \texttt{kaan.gorur@roketsan.com.tr}
}

\begin{document}
\maketitle
\begin{abstract}
Document parsing is now widely used in applications, such as large-scale document digitization, retrieval-augmented generation, and domain-specific pipelines in healthcare and education. Benchmarking these models is crucial for assessing their reliability and practical robustness. Existing benchmarks mostly target high-resource languages and provide limited coverage for low-resource settings, such as Turkish. Moreover, existing studies on Turkish document parsing lack a standardized benchmark that reflects real-world scenarios and document diversity. To address this gap, we introduce OCRTurk, a Turkish document parsing benchmark covering multiple layout elements and document categories at three difficulty levels. OCRTurk consists of 180 Turkish documents drawn from academic articles, theses, slide decks, and non-academic articles. We evaluate seven OCR models on OCRTurk using element-wise metrics. Across difficulty levels, PaddleOCR achieves the strongest overall results, leading most element-wise metrics except figures and attaining high Normalized Edit Distance scores in easy, medium, and hard subsets. We also observe performance variation by document type. Models perform well on non-academic documents, while slideshows become the most challenging.
\end{abstract}

\section{Introduction}

Optical Character Recognition (OCR) is a technology that enables the extraction of text, tables, figures, and other structural elements from images of handwriting \citep{memon2020handwrittenopticalcharacterrecognition}, receipts \citep{Huang_2019,park2019cord}, scenes \citep{Munjal_2021,Lunia_2023}, documents \citep{ouyang-etal-2024-omnidocbench}, and similar sources. Similarly, besides extracting raw text, extracting different elements is crucial for real-world applications. Table extraction \citep{Anand_2023,patel2025designimplementationocrpoweredpipeline,pallavi2020conglomeratemultipleocrtable} and equation extraction \citep{zhong2025doctronformulageneralizedformularecognition} are among the most popular element-wise extraction tasks in document understanding. This technology bridges the gap between image-based content and computer-based processing, making text accessible for downstream applications, including large language models (LLMs). 

Since the outputs of OCR models are used in daily life and as data for training machine learning systems, the reliability of these outputs is a very important issue. To address this and to observe model performance, several benchmarks are used \citep{ouyang-etal-2024-omnidocbench,poznanski-etal-OlmOCR,fu-etal-2024-ocrbench-v2}. These benchmarks measure the capability of models to extract text in the correct format and reading order, and to recognize tables, mathematical formulas, and figures called Document Parsing in total and understanding of key information of the documents. However, most existing benchmarks are in English, which makes it difficult to evaluate model performance in low-resource languages, such as Turkish. Turkish morphology, flexible word order, and variation across sources introduce challenges not reflected in English benchmarks \citep{hakkani2002statistical,oflazer2014turkish,umutlu-etal-2025-evaluating}.

From a Turkish perspective, although studies on text recognition increase over the years \citep{TurkishOCRSurvey}, there are still only a limited number of datasets and a single OCR Document Parsing benchmark in Turkish \citep{yilmaz-etal-2025-turkish-ocr-benchmark}. This benchmark evaluates model performance in Turkish character confusion, word length effects, context-dependent errors, and recognition under distortions, using synthetic Turkish data. However, this benchmark only evaluates the model’s performance on raw text, does not include elements, such as figures, tables, or mathematical equations, and uses synthetic data derived from a single source. For this reason, there is a clear gap in benchmarks that assess model performance beyond raw text and under realistic conditions.

In this work, to fill this gap, we introduce the first Turkish OCR benchmark\footnote{\url{https://github.com/metunlp/ocrturk}} designed to reflect real-world document diversity, difficulty, and structure. The benchmark contains 180 document pages in PDF format from articles, theses, non-academic documents, and slideshows. These documents come from various sources shared on GitHub and include tables, mathematical equations, and figures. These documents are divided into three difficulty levels: easy, medium, and hard based on the complexity of document structure.

Using this benchmark, it is possible to evaluate:
\paragraph{Raw text analysis:} The model's capability to recognize Turkish characters and extract raw text correctly.
\paragraph{Table recognition:} The model's ability to reproduce tables in the correct format and with accurate content.
\paragraph{Mathematical equation recognition:} The capability to reproduce formulas and equations from document images.
\paragraph{Figure recognition:} The capability to extract figures correctly from document images.

Compared with earlier Turkish benchmark, we make three main contributions. First, we move beyond raw-text–only evaluation and build a benchmark that reflects real-world documents from diverse sources, including figures, tables, and equations. Second, we publicly release the dataset and evaluation scripts so that researchers can test their models on many practical scenarios before using them on real documents. We show that model performance differs across document types and difficulty levels. PaddleOCR achieves the best overall results. In addition, models perform the best on non-academic documents and the worst on slideshows.

\section{Related Work}

\paragraph{Document Parsing Benchmarks}
Document parsing is an important aspect of OCR, as it measures the capability of models to extract different structural elements from documents. OmniDocBench \citep{ouyang-etal-2024-omnidocbench} is a benchmark that contains 981 PDF pages across nine distinct document types and evaluates raw text correction, table recognition, formula recognition, and reading order. It also uses adjacency-search matching to reduce the impact of paragraph splitting and an ignore-handling strategy to exclude parts of the text, such as headers and footers, in order to obtain more consistent metrics. Similarly, OlmOCR-Bench \citep{poznanski-etal-OlmOCR} is a widely used benchmark. Like OmniDocBench, it measures text presence and absence, natural reading order, table accuracy, and mathematical formula accuracy. It includes a total of 7{,}010 test PDF pages across these categories. All tests follow a pass/fail format, and the overall score is computed as the mean over all test categories. The CC-OCR \citep{yang-etal-2024-cc-ocr} benchmark is designed for different real-world scenarios. It includes tasks for multi-scene OCR, multilingual OCR, document parsing, and key information recognition. KITAB \citep{heakl2025kitabbenchcomprehensivemultidomainbenchmark} is an Arabic OCR benchmark with 8,809 samples across nine domains. It includes tasks, such as table recognition and PDF-to-Markdown conversion, and it provides a strong example of an OCR benchmark for a low-resource language. In real-world document parsing, some documents contain photographs taken from different angles, which earlier benchmarks do not explicitly address. DocPTBench \citep{du-etal-2025-docptbench} is introduced to fill this gap and evaluates document parsing abilities for various real-world cases, including digital-born documents, photographed documents, and unwrapped photographed documents.
\paragraph{Turkish Document Parsing}
Turkish document parsing studies are mainly grouped into two categories: Handwriting parsing and digital character parsing. In handwriting studies, some work uses Turkish handwritten characters that are created in computer environments \citep{kuncan-etal-2020-turkish-handwriting,al-zubaidi-etal-2019-two-dimensional-ocr}. In these studies, characters are generated by using a mouse on a computer, and the handwritten characters are classified using Artificial Neural Networks (ANNs). \citet{kizilirmak-2022-offline-handwriting} studies the same problem using Convolutional Neural Networks (CNNs) and BiLSTM architectures, and also introduces a new dataset that contains 2{,}600 neutral handwriting line images collected from 73 participants. On the other hand, \citet{sevik-2019-derin-ogrenme} studies Turkish character recognition in digital texts with different fonts. This work uses 13{,}000 Turkish letters with 38 different fonts, and classifies them with CNN-based models. Similarly, there are several studies that focus on non-handwritten documents. NacsoftOCR \citep{Sayallar2023} is an example of Turkish receipt recognition, while \citet{b20eb0cbcfa7411d83518db7a94b879a} studies information extraction from bank documents, such as money transfer orders.
\paragraph{Turkish Document Parsing Datasets and Benchmarks}

There are Turkish LLM benchmarking efforts that involve RAG evaluation \citep{toraman2026turkbenchbenchmarkevaluatingturkish}. However, they are limited in this particular task.

THE dataset \citep{bartos-etal-2020-the-dataset} is a multilingual handwritten character dataset that contains Turkish, Hungarian, and English characters. The dataset is collected from 200 participants and includes 15,600 binary character images corresponding to 78 unique letters.

Türkçe Kitap \citep{zeer2024cosmos} is a dataset constructed from a collection of 100,000 books. It contains images taken from these books, as well as conversations between large language models and humans about the visual content.

To the best of our knowledge, the only benchmark in Turkish for OCR and vision-language models is the work by \citet{yilmaz-etal-2025-turkish-ocr-benchmark}. They introduce a synthetic Turkish dataset that includes 6,600 images with printed, handwritten, and scene text. They use Mustafa Kemal Atatürk’s book, Nutuk, as the text source, and place selected words and sentences onto background images taken from the COCO dataset \citep{lin-etal-2014-microsoft-coco} after removing all textual content from the images for the scene-text setting. They use the Text Recognition Data Generator\footnote{https://github.com/Belval/TextRecognitionDataGenerator} to create both handwritten and printed text. They evaluate the models with respect to Turkish character confusion, word-length effects, context-dependent errors, and distortion type, and they use Character Error Rate (CER) and Word Error Rate (WER) as evaluation metrics.

\section{Dataset}
\paragraph{Categorization of Data} We categorize the dataset into four categories: academic documents, non-academic documents, theses, slideshows. While academic documents are only from arxiv\footnote{https://arxiv.org/} and DergiPark\footnote{https://dergipark.org.tr/tr/} non-academic documents include financial reports, course materials, manuals, and annual summary reports from various sources (which are available at the JSON files of each data in the dataset). All theses are from YÖK TEZ\footnote{https://tez.yok.gov.tr/UlusalTezMerkezi/} and the slideshows are from MEB OGM Materyal\footnote{https://ogmmateryal.eba.gov.tr/} and Ankara Üniversitesi Açık Öğretim Materyali\footnote{https://acikders.ankara.edu.tr/}. We generate equal amounts of data for each category (45 pages of data for each category).
\paragraph{Classification of Data} We classify every data in the dataset as one of the three difficulty levels: easy, medium, and hard. We classify the documents that include only texts as easy. We classify the documents as medium if they include both texts and one of the following items once: one-line equation, tables without multi-columns or multi-rows, figures without sub-figures. Otherwise, we classify the data as hard. We generate equal amounts of data for each difficulty (60 pages of data for each difficulty).
\paragraph{Data Structure} Observing that the OCR models return the outputs in Markdown format, we adopt this format in our dataset to standardize the data gathered from various sources. This allows us to use HTML format for the tables, LaTeX format for the equations, and PNG format for the figures. To generate statistics for the dataset, we store the tables, equations, and images in separate folders for each data.

\paragraph{Data Generation} The benchmark is constructed from Turkish documents collected from multiple sources. The dataset is split into two disjoint subsets. Two annotators each converts one subset of the original documents into a unified Markdown representation. Prior to conversion, repetitive headers and footers are cropped from the original PDFs. This is done because a large portion of models do not take them into account while generating responses, and to ensure that the strengths of each model are evaluated fairly. This method is also adopted in \citep{ouyang-etal-2024-omnidocbench}. During this conversion, plain text is transcribed verbatim, tables are converted into HTML format, and mathematical expressions are converted into LaTeX. ChatGPT and Gemini are used as assistive tools during this conversion step to accelerate formatting, but all outputs are manually verified.

Following the conversion, each annotator performs a character-level manual verification of their own converted Markdown files against the original documents. This includes checking textual content, table structure, and mathematical expressions to ensure faithful transcription. Discrepancies were corrected manually.

As a quality control step, only samples that passes manual character-level consistency checks are included in the final benchmark. This manual checking step ensures strong consistency between the source documents and the benchmark annotations. Since the data are manually generated and carefully inspected character by character, the data construction process is highly time-consuming; as a result, the benchmark currently consists of 180 pages of data.

\paragraph{Summary} The number of pages of data in each category and difficulty level and their totals are explained in Table~\ref{tab:summary_of_the_dataset}. The total number of items (equations, tables, and figures) in each category and in each difficulty level and their respective totals are explained in Table~\ref{tab:stats_wrt_categories_and_difficulty_levels}.

\begin{table}[t]
  \centering
  \resizebox{\columnwidth}{!}{%
      \begin{tabular}{lrrrc}
        \toprule
        \multirow{2}{*}{\textbf{Category}} & \multicolumn{3}{c}{\textbf{Difficulty}} & \multirow{2}{*}{\textbf{Total}} \\
        \cline{2-4}
        & \multicolumn{1}{l}{Easy} & \multicolumn{1}{l}{Medium} & \multicolumn{1}{l}{Hard} & \\
        \midrule
        \multicolumn{1}{l}{\textbf{Academic Docs.}} & 15 & 15 & 15 & \multicolumn{1}{r}{45}\\
        \multicolumn{1}{l}{\textbf{Non-academic Docs.}} & 15 & 15 & 15 & \multicolumn{1}{r}{45}\\
        \multicolumn{1}{l}{\textbf{Theses}} & 15 & 15 & 15 & \multicolumn{1}{r}{45}\\
        \multicolumn{1}{l}{\textbf{Slideshows}} & 15 & 15 & 15 & \multicolumn{1}{r}{45}\\
        \midrule
        \multicolumn{1}{l}{\textbf{Total}} & 60 & 60 & 60 & \multicolumn{1}{r}{\textbf{180}}\\
        \bottomrule
      \end{tabular}
    }
  \caption{\label{tab:summary_of_the_dataset}
    The table summarizes the number of pages of data for each category and difficulty level. The totals in each category and in each difficulty level is given. The subtotal is given in the bottom-right corner in bold.
  }
\end{table}

\begin{table}[t]
  \centering
  \resizebox{\columnwidth}{!}{%
    \begin{tabular}{clrrrr}
      \toprule
      \multicolumn{2}{c}{\multirow{2}{*}{\phantom{abc}}} & \multicolumn{3}{c}{\textbf{Items}} & \multirow{2}{*}{\textbf{Total}} \\
      \cline{3-5}
      \multicolumn{2}{c}{} & Equations & Tables & Figures & \\
      \midrule
      
      \multirow{5}{*}{\rotatebox[origin=c]{90}{\textbf{Category}}} 
      & \textbf{Academic Docs.} & 42 & 42 & 22 & 106 \\
      & \textbf{Non-academic Docs.} & 0 & 45 & 16 & 61 \\
      & \textbf{Theses} & 42 & 7 & 16 & 65 \\
      & \textbf{Slideshows} & 8 & 36 & 3 & 47 \\
      \cline{2-6}
      & \textbf{Total} & \textbf{92} & \textbf{130} & \textbf{57} & \textbf{279} \\
      \midrule

      \multirow{4}{*}{\rotatebox[origin=c]{90}{\textbf{Difficulty}}} 
      & \textbf{Easy} & 0 & 0 & 0 & 0 \\
      & \textbf{Medium} & 11 & 51 & 23 & 85 \\
      & \textbf{Hard} & 81 & 79 & 34 & 194 \\
      \cline{2-6}
      & \textbf{Total} & \textbf{92} & \textbf{130} & \textbf{57} & \textbf{279} \\
      \bottomrule
    \end{tabular}
  }
  \caption{\label{tab:stats_wrt_categories_and_difficulty_levels}
    The table summarizes the total number of items (equations, tables, and figures) in each category and for each difficulty level.}
\end{table}

\section{Methodology}
We evaluate the model accuracy based on how correctly they reproduce the texts, tables, equations, and figures in the given documents. To do this, we first extract the items (equations, tables, and figure tags) from the Markdown, to be used for the evaluations metrics afterwards, and end up with raw texts. Then we apply a number of post processing steps on the raw texts. We convert the misprinted Turkish characters. For example, we convert $\Breve{}$ g (breve symbol followed by the letter g) to ğ, and \c{} s (cedilla symbol followed by the letter s) to ş. We remove the title, subtitle, etc. tags (\#, \#\#, **). After these post-processing steps, we continue with the calculation of the performance scores with the evaluation metrics using the extracted items and cleaned raw texts.

\subsection{Evaluation Metrics}

\subsubsection{Texts}

For texts, we use two metrics: Normalized edit distance (NED) and Turkish character sensitivity (TCS). In \emph{normalized edit distance}, we calculate the number of edits that we should apply to the model output to achieve the same result as the ground truth. Formally defined as
\[
    \text{NED} = \frac{d_{edit}(H,R)}{max\{|H|, |R|\}}
\]
where $d_{edit}(H,R)$ shows the Levenshtein Distance between the model output $H$ and the ground truth $R$. To obtain a number between 0 and 1 for the score, we divide this distance by the maximum of the number of characters in the model output and in the ground truth. While lower scores represent a higher similarity between the model output and the ground truth, higher scores indicates that a number of edits should be applied to the model output to achieve the same result as the ground truth. This metric is used in OmniDocBench. In \emph{Turkish character sensitivity}, we try to measure how accurately models reproduce characters specific to Turkish (ç, ğ, ı, ö, ş, ü, Ç, Ğ, İ, Ö, Ş, Ü). To achieve this, we calculate the ratio of errors to the total number of characters specific to Turkish. Formally defined as
\[
    \text{TCS} = 1 - \frac{E}{N}
\]
where $E$ and $N$ represents the number of errors and the total number of characters specific to Turkish, respectively. We subtract this index from 1 to obtain a score where a higher value indicates a better performance (e.g. 0 errors, $E = 0$, will return a score of 1). 

\subsubsection{Tables}
For tables, we use two metrics: Tree edit distance based similarity (TEDS) and normalized edit distance (NED). In \emph{Tree edit distance based similarity}, we first convert the tables to tree structures, then calculate the number of edits that we should apply to the tree generated from model output to achieve the same result as the tree generated from the ground truth. Formally defined as
\[
    \text{TEDS} = 1 - \frac{d_{edit}(T_H,T_R)}{max\{|T_H|, |T_R|\}}
\]
where $d_{edit}(T_H,T_R)$ shows the tree edit distance between the tree generated from the model output $T_H$ and the tree generated from the ground truth. Similarly to NED, we divide this by the maximum of the number of nodes in the trees generated from the model output and the ground truth. Since we measure the similarity score, we subtract this fraction from 1. Thus, higher scores represent a higher similarity between the tables. In \emph{normalized edit distance}, we adopt a similar approach to the NED for texts. The difference is that for this metric, we consider the content of the table, such as cell values and table tags. Lower scores in this metric represent a higher similarity.

\subsubsection{Equations}
For equations, we use three metrics: Bilingual evaluation understudy score (BLEU), character detection matching (CDM), and normalized edit distance (NED). In \emph{Bilingual evaluation understudy score}, we measure the overlap between the model output equation and the ground truth equation, both of which are in LaTeX format. Essentially, it's an $n$-gram based metric that evaluates the ratio of the symbols produced in the correct order. Formally defined as
\[
    \text{BLEU} = \text{BP} \cdot \exp\Bigg({\sum_{n=1}^{N}{w_n\log{p_n}}}\Bigg)
\]
where BP penalizes the model output for being shorter than the ground truth, brevity penalty. $p_n$, $w_n$, and $N$ refers to the $n$-gram precision, weights for every $n$-gram (usually $w_n = 1/N$, and the maximum $n$-gram length (e.g. $N = 4$), respectively. Since the ordering of symbols is critical for correct equation interpretation, BLEU is used to assess the model’s ability to generate syntactically correct expressions. This metric is used in OmniDocBench. In \emph{normalized edit distance}, we adopt a similar approach to the NED for texts. While NED measures the structural similarity of the equations, it also takes the minor symbol differences into account. It's used to detect the character level errors especially for long and/or symbolically crowded equations (e.g. integrals, matrices). In \emph{character detection matching}, we measure how precisely the models recognize the characters in equations. True positives, false negatives, and false positives are evaluated together. Formally defined as
\[
    \text{CDM} = \frac{2 \cdot \text{TP}}{2 \cdot \text{TP} + \text{FP} + \text{FN}}
\]
where TP, FN, and FP represent true positives, false negatives, and false positives, respectively. It reflects the performance drop of models precisely especially when the models misinterpret the symbols like subscripts ($_\text{2}$), superscripts ($^\text{2}$), and special symbols ($\sigma$). This metric is used in OmniDocBench.

\subsubsection{Figures}
For figures, we use two metrics: Mean squared error (MSE), and DreamSim's evaluation score (DS) \citep{fu2023dreamsimlearningnewdimensions}. We use \emph{mean squared error} to quantify the average squared difference between the model output and the ground truth image. Formally defined as
\[
    \text{MSE} = \frac{1}{N}\sum_{i = 1}^{N}{(R'_{i} - H'_{i})^2}
\]
where $N$ is the total number of elements (pixels $\times$ channels) in the images. $H'_{i}$ and $R'_{i}$ represent the ${i}^\text{th}$ element in the images $H'$ and $R'$, respectively. We obtain the images $H'$ and $R'$ by normalizing the pixel values of the original model output $H$ and the ground truth $R$ to $[0,1]$ range. Lower MSE scores indicate lower average squared differences between the model output and the ground truth image, showing a higher similarity between them. We adopt the \emph{DreamSim evaluation metric}, a deep learning-based similarity metric, to evaluate perceptual quality, as recent studies indicate that it correlates highly with human visual assessment and is robust to cropped images compared to other metrics. \citep{wickrema2025benchmarkingimagesimilaritymetrics}.

\subsection{Experimental Setup}
In this paper, we evaluate OlmOCR2\footnote{https://huggingface.co/allenai/OlmOCR2-2-7B-1025}, DeepSeek-OCR\footnote{https://github.com/deepseek-ai/DeepSeek-OCR}, NanonetsOCR2\footnote{https://huggingface.co/nanonets/Nanonets-OCR2-3B}, PaddleOCR\footnote{https://github.com/PaddlePaddle/PaddleOCR}, Docling\footnote{https://github.com/docling-project/docling}, and NVIDIA Nemotron v1.1\footnote{https://huggingface.co/nvidia/NVIDIA-Nemotron-Parse-v1.1}, HuanyanOCR\footnote{https://huggingface.co/tencent/HunyuanOCR} Parse using the proposed benchmark.

\paragraph{DeepSeekOCR}
We run DeepSeekOCR inference using the Transformers model.infer method, where the base size is set to 1,024 and the image size is set to 640. The crop mode and test compression options are enabled, while all other inference parameters are kept at their default values.
\paragraph{DoclingOCR}
For DoclingOCR, we use Docling’s official library and perform document conversion through the DocumentConverter, specifying PDF processing via the PdfFormatOption with customized pipeline options. In particular, the image scale is set to 2, and both page image generation and picture image generation are enabled. All other pipeline and conversion parameters are kept at their default values.

\paragraph{PaddleOCR}
For PaddleOCR, we use the official PaddlePaddle OCR library and perform inference with the {PaddleOCRVL} using its default configuration.

\paragraph{HuanyanOCR}
For HuanyanOCR, we perform text generation using the Hugging Face Transformers library, where the maximum number of new tokens is set to 16,384 and sampling is disabled. All other generation parameters are kept at their default Transformers settings.

\paragraph{Nanonets OCR2}
For Nanonets OCR, we perform text generation using the Hugging Face Transformers library, where the maximum number of new tokens is set to 15,000 and sampling is disabled. All other generation parameters are left at their default Transformers values.

\paragraph{Nvidia Nemotron v1.1}
For NVIDIA OCR, we perform text generation using the Hugging Face Transformers library with a customized configuration. Specifically, the beginning-of-sequence token identifier is set to 0, the decoder start token identifier and end-of-sequence token identifier are both set to 2, the forced end-of-sequence token identifier is set to 2, and the padding token identifier is set to 1. In addition, the maximum number of new tokens is set to 9,000, sampling is disabled, beam search is performed with a single beam, and a repetition penalty of 1.1 is applied. All other generation parameters are left at their default Transformers values.

\paragraph{OlmOCR2}
For OlmOCR2, we perform text generation using the Hugging Face Transformers library, where the temperature is set to 0.1, the maximum number of new tokens is set to 15,000, a single output sequence is generated, and sampling is enabled. All other generation parameters are kept at their default Transformers settings.

All model parameters are adopted from the official example usages provided in the models’ Hugging Face pages or GitHub repositories. This choice is motivated by the goal of following developer-recommended configurations to obtain the best possible outputs under standard settings. All outputs are generated in the medium of L4 22.5 GB GPU and A100 40 GB GPU.

\begin{table*}[t]
\centering
\setlength{\tabcolsep}{4pt}
    \begin{tabular}{l|rr|rrr|rr|rr}
    \toprule
    \multirow{2}{*}{\textbf{OCR Model}} & \multicolumn{2}{c|}{\textbf{Raw Texts}} & \multicolumn{3}{c|}{\textbf{Equations}} & \multicolumn{2}{c|}{\textbf{Tables}} & \multicolumn{2}{c}{\textbf{Figures}} \\
    \cline{2-10}
    & \multicolumn{1}{l}{NED $\downarrow$} & \multicolumn{1}{l|}{TCS $\uparrow$} & \multicolumn{1}{l}{NED $\downarrow$} & \multicolumn{1}{l}{BLEU $\uparrow$} & \multicolumn{1}{l|}{CDM $\uparrow$} & \multicolumn{1}{l}{NED $\downarrow$} & \multicolumn{1}{l|}{TEDS $\uparrow$} & \multicolumn{1}{l}{MSE $\downarrow$} & \multicolumn{1}{l}{DS $\downarrow$} \\ 
    \midrule
    \multicolumn{1}{l|}{DeepSeekOCR} & 0.12 & 0.81 & 0.08 & 0.91 & 0.92 & \textbf{0.05} & \textbf{0.87} & \textbf{0.04} & \textbf{0.06} \\ 
    \multicolumn{1}{l|}{Docling}  & 0.13 & 0.71 & 0.15 & 0.85 & 0.85 & \textbf{0.05} & 0.86 & 0.06 & 0.12 \\ 
    \multicolumn{1}{l|}{PaddleOCR}   & \textbf{0.08} & 0.82 & 0.06 & \textbf{0.94} & \textbf{0.95} & \textbf{0.05} & \textbf{0.87} & 0.05 & 0.10 \\
    \multicolumn{1}{l|}{HuanyanOCR}  & 0.09 & \textbf{0.88} & 0.11 & 0.89 & 0.89 & 0.06 & 0.85 & - & -    \\ 
    \multicolumn{1}{l|}{NanonetsOCR2} & 0.17 & 0.79 & \textbf{0.05} & \textbf{0.94} & \textbf{0.95} & 0.11 & 0.80 & - & -    \\ 
    \multicolumn{1}{l|}{Nvidia Nemotron v1.1} & 0.27 & 0.47 & 0.14 & 0.85 & 0.86 & 0.07 & 0.86 & - & -    \\ 
    \multicolumn{1}{l|}{OlmOCR2} & 0.09 & 0.80 & 0.12 & 0.87 & 0.88 & \textbf{0.05} & 0.85 & - & -    \\
    \bottomrule
    \end{tabular}
    \caption{\label{tab:quantitative_results_complete_table}
    The summary of the average scores of the models DeepSeekOCR, Docling, PaddleOCR, HuanyanOCR, NanonetsOCR2, Nvidia Nemotron v1.1, and OlmOCR using the evaluation metrics NED (normalized edit distance), TCS (Turkish character sensitivity), BLEU (bilingual evaluation understudy), CDM (character detection matching), TEDS (tree edit distance based similarity), MSE (mean squared error), and DS (DreamSim's evaluation score). The upward arrows ($\uparrow$) next to the metric abbreviations indicate a better performance if the value is higher, while the downward arrows ($\downarrow$) indicate a better performance if the value is lower. The best result(s) within each metric is written in bold, accordingly. We indicate the values for the image metric columns with dashes if the models do not return images in their outputs.
    }
\end{table*}

\section{Results}
After obtaining the scores of the OCR models for each evaluation metric within 180 pages of data, we calculate metrics of these with all data. We give the detailed results in Table~\ref{tab:quantitative_results_complete_table}. 

\subsection{Element Based Results}
\paragraph{Raw Texts} PaddleOCR, HuanyanOCR, and OlmOCR2 lead the raw-text evaluation with HuanyanOCR achieving the highest TCS score, indicating that it performed best with Turkish charaters. They achieve low Normalized Edit Distance  and high Turkish Character Sensitivity , indicating that their outputs require minimal edits and that they better preserve Turkish-specific characters. While Docling attains a comparable NED, its lower TCS suggests weaknesses in handling Turkish characters. Nvidia Nemotron v1.1 performs worst in this section (NED = 0.27, TCS = 0.47), indicating substantial errors and limited sensitivity to Turkish characters.

\paragraph{Equations} PaddleOCR remains the strongest on the equation benchmark, consistent with its raw-text performance. NanonetsOCR2 joins the top group and leads overall, achieving the best scores across all three metrics (NED = 0.05, BLEU = 0.94, CDM = 0.95). Although DeepSeekOCR falls behind PaddelOCR and NanonetsOCR2, it shows a relatively good performance. The remaining models cluster in a mid-range performance band, with NED in the 0.11–0.15 range and BLEU/CDM between 0.85 and 0.89, indicating solid equation transcription quality but a clear gap to the leaders.

\paragraph{Tables} Most models perform very well on table extraction, with NED values around 0.05 and TEDS scores close to 0.87, that shows model has capability to extract both contents and the tree level structure correctly. NanonetsOCR2 is the main exception, showing lower performance (NED = 0.11, TEDS = 0.80).

\paragraph{Figures} Since only DeepSeekOCR, Docling, and PaddleOCR are able to extract figures from the documents, we evaluate only these models in the figure section. DeepSeekOCR leads on figure extraction, achieving the lowest MSE (0.04) and the lowest DreamSim score (DS = 0.06). All three models obtain very low MSE values, indicating that their extracted figures are highly similar to the ground truth at the pixel level. However, PaddleOCR and Docling show high DreamSim scores, suggesting lower perceptual similarities despite comparable pixel-level accuracies.

\subsection{Difficulty Level Based Results}
\begin{figure*}[t]
  \centering
  \includegraphics[width=0.6\linewidth]{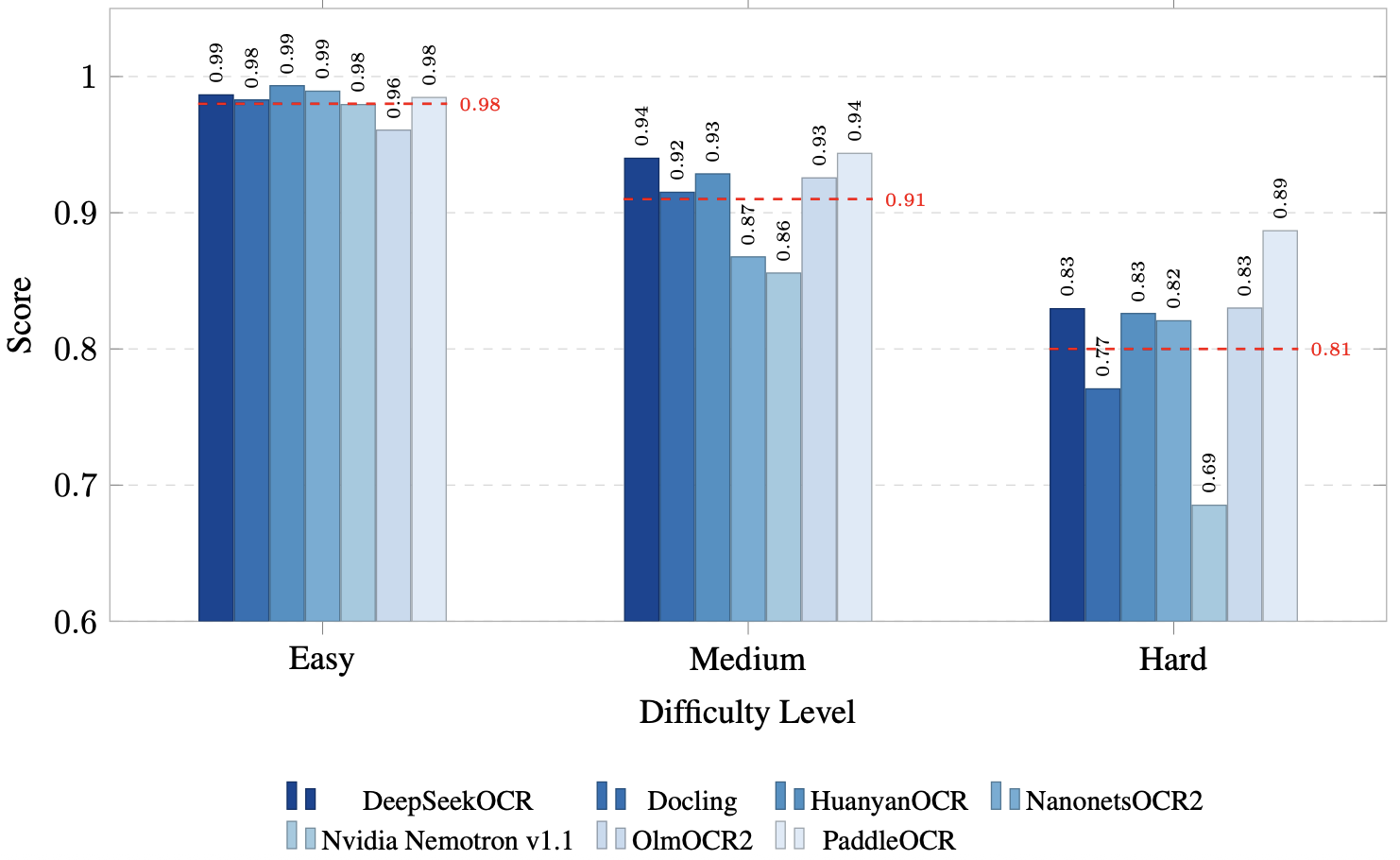}
  \caption{Comparison of the models DeepSeekOCR, Docling, PaddleOCR, HuanyanOCR, NanonetsOCR2, Nvidia Nemotron v1.1, and OlmOCR2 under easy, medium, and hard data. The scores are the averages of the NED metric scores for raw texts, tables, and equations of the data within the same difficulty. The NED scores are subtracted from 1 ($\text{Score} = 1 - \text{NED}$) for better comparison. The average score of the models within each difficulty level is given as the dashed red line.}
  \label{fig:metric_comparison_wrt_difficulty}
\end{figure*}

Difficulty-based results are summarized in Figure~\ref{fig:metric_comparison_wrt_difficulty}. For each model, we first compute the average NED across raw text, tables, and equations, and then report a unified score as \(1 - \mathrm{avg}(\mathrm{NED})\), so that higher values indicate better performance and comparisons are more intuitive.

In the easy category, all models achieve near-perfect performance, with an average score of 0.98. In the medium category, performance remains high with an average score of 0.91. DeepSeekOCR, Docling, HuanyanOCR, OlmOCR2, and PaddleOCR perform above the overall average, even though the documents include basic structure, such as simple tables and short equations. In the hard category, PaddleOCR achieves the best score (0.89), despite the presence of more complex content, such as multi-line equations, plots with subplots, and larger tables. Overall, 5 out of 7 models score above the average in the hard set, while Docling and Nvidia Nemotron v1.1 fall below the average.

\subsection{Category Based Results}

\begin{figure*}[t]
  \centering
  \includegraphics[width=0.6\textwidth]{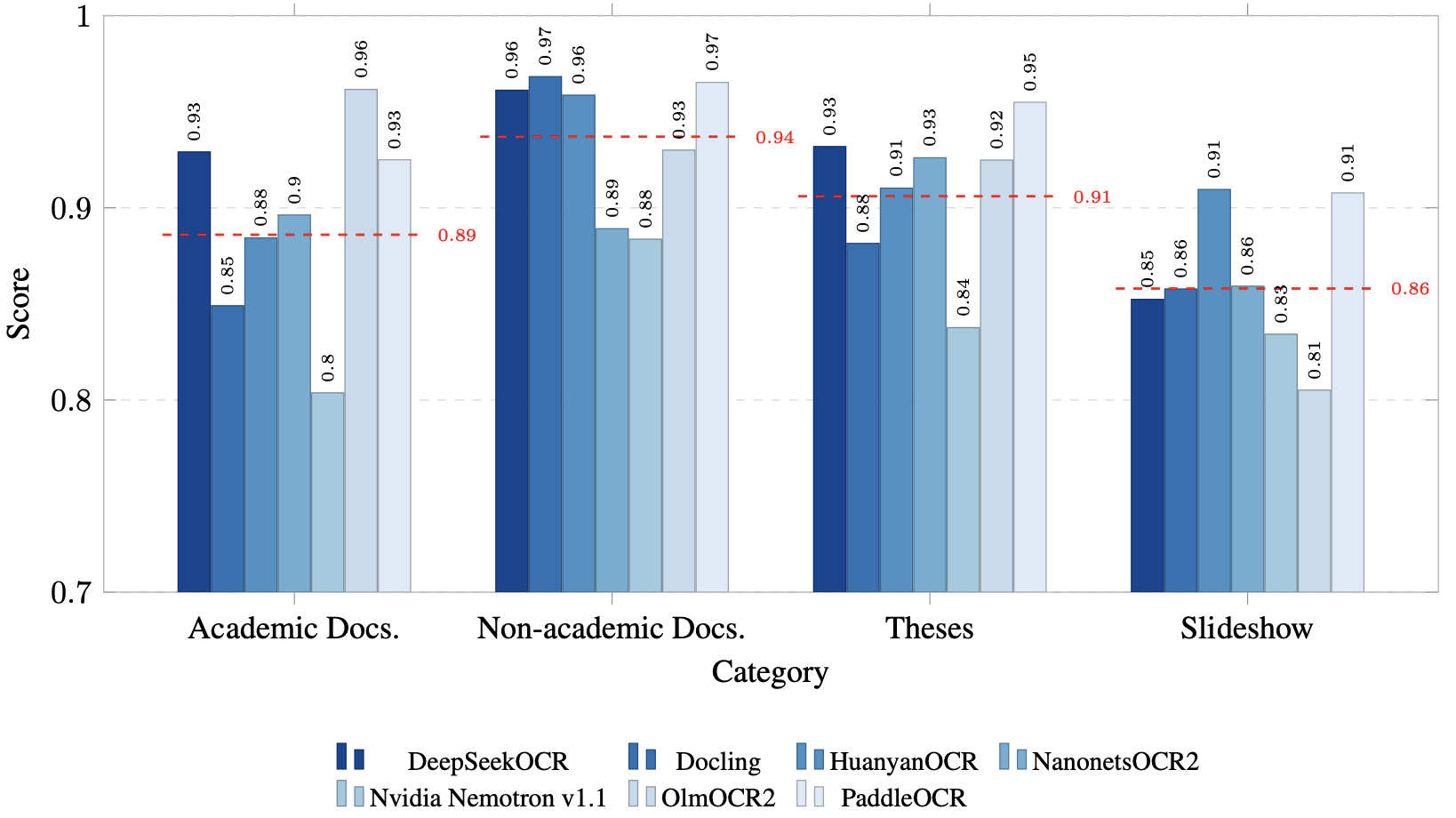}
  \caption{Comparison of the models DeepSeekOCR, Docling, PaddleOCR, HuanyanOCR, NanonetsOCR2, Nvidia Nemotron v1.1, and OlmOCR2 under the categories academic documents, non-academic documents, theses, and slideshows. The scores are the averages of the NED metric scores for raw texts, tables, and equations of the data within the same difficulty. The NED scores are subtracted from 1 ($\text{Score} = 1 - \text{NED}$) for better comparison. The average score of the models within each category is given as the dashed red line.}
  \label{fig:metric_comparison_wrt_category}
\end{figure*}

Category-based results are summarized in Figure~\ref{fig:metric_comparison_wrt_category}. For each model, we first compute the average NED across raw texts, tables, and equations. We then report a unified score as $1 - \text{avg(NED)}$, so that higher values indicate better performance and comparisons are more intuitive.

Among the four categories (academic documents, non-academic documents, theses, and slideshows), non-academic documents show the highest average score with $0.94$ whereas slideshows show the lowest average score with $0.86$. Among the seven models (DeepSeekOCR, Docling, PaddleOCR, HuanyanOCR, NanonetsOCR2, Nvidia Nemotron v1.1, and OlmOCR2), PaddleOCR either has the highest or tied-highest scores in three of the four categories. Nvidia Nemotron v1.1 shows the lowest scores in academic documents, non-academic documents, and theses. While HuanyanOCR and PaddleOCR performed the best, OlmOCR2 performed the worst in slideshows. For Docling, we observe the highest gap ($0.12$) between the two categories: academic docs ($0.85$) and non-academic docs ($0.97$). Overall, only PaddleOCR performed above the average in all four categories. The observed performance gap between slideshow documents and other document types can be attributed to the inherently unstructured nature of slideshows. Unlike theses and academic papers, which typically follow well-defined formatting standards, slideshows often contain highly unstructured text, irregular layouts, and ambiguous or incomplete table representations. This lack of standardized structure increases extraction difficulty and leads to degraded performance compared to more consistently formatted documents.

\subsection{Error Analysis}
The models share recurring errors when extracting Turkish special characters, leading to missing titles, line skips and line-break issues. They produce incorrect variables in equations (such as range shifts in integrals or sums) and sometimes write the equations or mathematical variables in text mode instead of in math mode, resulting in an incorrect number of equations in the document. The models sometimes skip the tables or fail to recognize them as tables, treating them as text instead. They occasionally generate data for empty cells or create extra columns. In cases where a column of the document includes an image, the models often interpret the entire section as an image. All of these errors combined results in penalties to these models' accuracy scores. A detailed error analysis with examples is provided in \autoref{sec:appendix_a}.

\section{Conclusion}
Prior Turkish document parsing research highlights a clear gap: the absence of a Turkish document parsing benchmark that reflects real-world scenarios. To address this need, we introduce OCRTurk, the first, to the best of our knowledge, Turkish document parsing benchmark. OCRTurk comprises 180 documents spanning multiple document types and three difficulty levels, designed to capture the diversity and challenges of real-world Turkish documents. We evaluate seven OCR models on OCRTurk.

Most of the models show their strength for different element types, such as NanonetsOCR2 performing the best for equation, but the worst for tables. However, PaddleOCR achieves the strongest overall performance on element-level metrics, while DeepSeekOCR leads in figure extraction and HuanyanOCR leads in Turkish Charater Sensitivity. Under difficulty-based evaluation, DeepSeekOCR, HuanyanOCR, and PaddleOCR consistently outperform the overall average across all difficulty levels. Performance also varies substantially by document type: models perform best on non-academic documents and theses, and worst on slideshows.

Future work will focus on expanding the dataset by incorporating additional samples and a broader range of document types. In parallel, we plan to release the benchmark through an online leaderboard, enabling systematic evaluation and comparison of newly developed models.

\section*{Limitations}
This benchmark contains 180 documents. Because generating and manually correcting ground truth annotations is time-consuming, we limit the benchmark to 180 documents in this release. Expanding the dataset would increase its diversity and enable evaluation on a wider range of edge-case scenarios. Similarly, extending the set of document types would better capture the variety of real-world documents and further strengthen the benchmark’s coverage.
\section*{Ethical Considerations}

OCRTurk is constructed using publicly accessible documents from platforms, such as ArXiv, DergiPark, YÖKTez, Ankara Üniversitesi Açık Kaynak Materyal, MEB OGM Materyal, and similar websites. The data are collected from a wide range of domains, including artificial intelligence, geography, finance, mathematics, and others, to reduce domain bias and evaluate models across diverse document types and writing styles. All data are collected in compliance with the terms of use of these platforms, and no private, sensitive, or restricted information is included. The dataset is intended strictly for academic research and for evaluating the performance of OCR and document parsing models; it has no commercial or financial objectives. Since the documents are already publicly available, the collection process does not violate individual privacy rights.

Regarding data annotation, Large Language Models, including GPT-5 and Gemini 2.5 Flash, are utilized to generate initial drafts of LaTeX and HTML structures. Recognizing the risk of model hallucinations, every annotation is manually reviewed and corrected by the authors to ensure the ground truth’s absolute accuracy. Additionally, LLMs are used solely for proofreading and refining the grammatical clarity of the human-written text to improve readability.

Generative AI (ChatGPT\footnote{\url{https://chatgpt.com/}} and Gemini\footnote{\url{https://gemini.google.com}}) is used in writing of this study to assist with language editing. All scientific contributions, data construction, data analysis, and interpretations presented in this work are original and were conducted entirely by the authors.

\section*{Acknowledgments}
We gratefully acknowledge the support of Google Academic Program for providing Google Cloud\footnote{\url{https://cloud.google.com/}} credits that facilitated this research. We express our sincere gratitude to Roketsan Inc.\footnote{\url{http://www.roketsan.com.tr/}} for their valuable support throughout the development of this work.

\bibliography{main}
\begin{figure*}[t]
    \centering
    \begin{subfigure}[b]{0.45\textwidth}
        \centering
        \includegraphics[width=\linewidth]{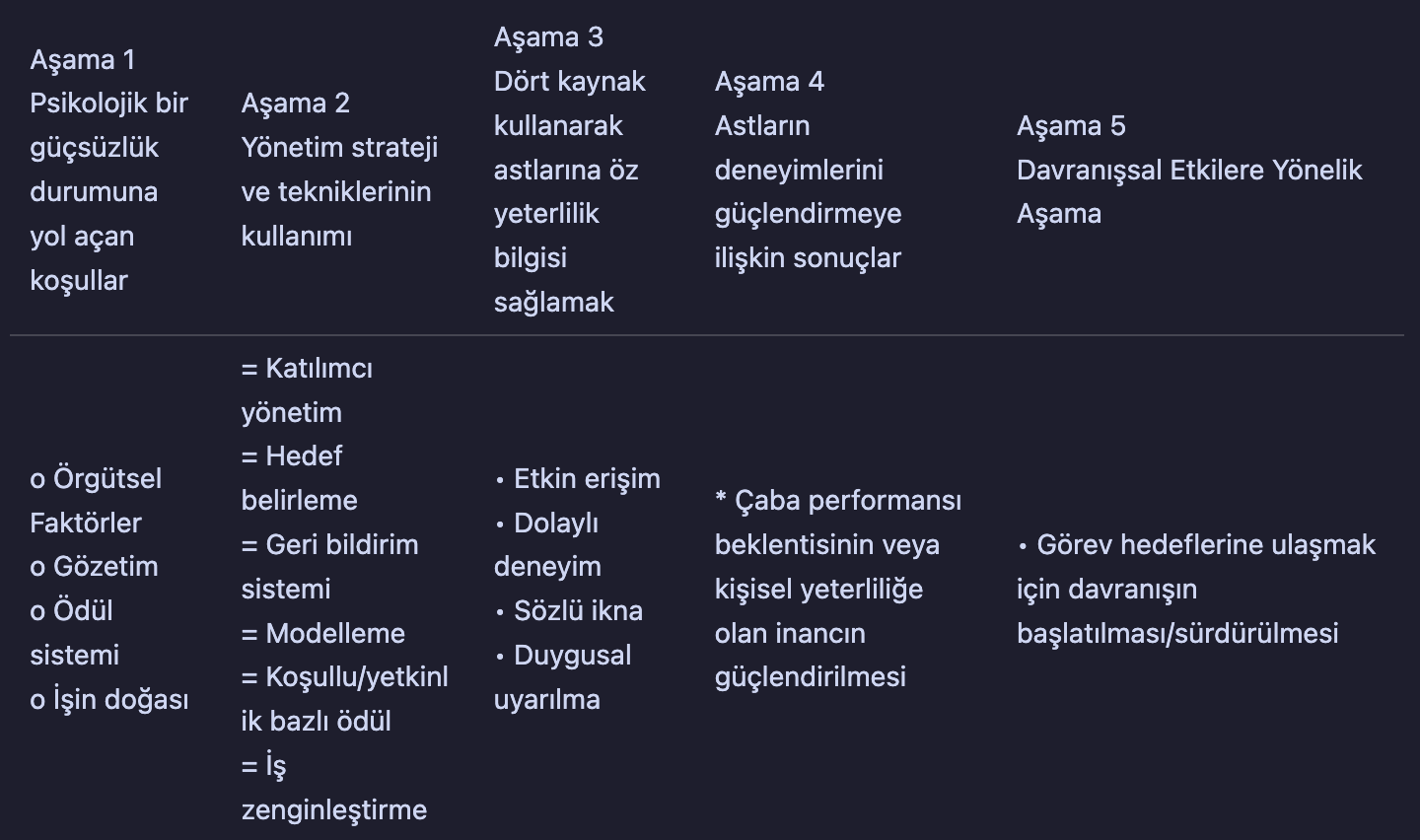}
        \caption{Ground Truth}
        \label{fig:ftable_errors_figure_image_1}
    \end{subfigure}
    \hfill 
    \begin{subfigure}[b]{0.45\textwidth}
        \centering
        \includegraphics[width=\linewidth]{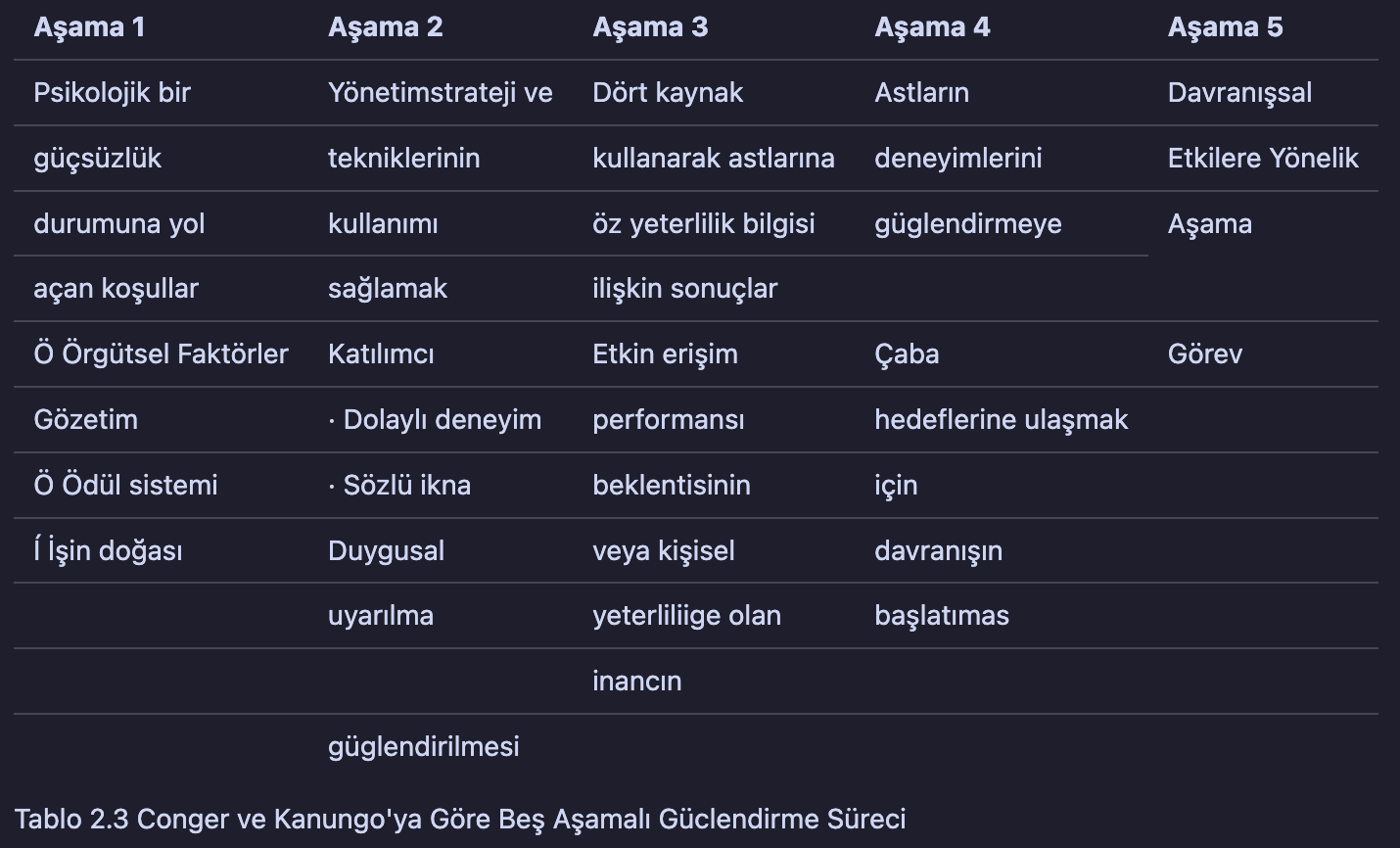}
        \caption{Model Output}
        \label{fig:table_errors_figure_image_2}
    \end{subfigure}
    
    \caption{Comparison of the tables between the ground truth (a) and the model output (b). As the structure of the table in the model output is incorrect, it spans to more than 50 lines. Thus, we ignore the rest of the model output in this figure.}
    \label{fig:table_errors_figure}
\end{figure*}
\begin{figure*}[t]
    \centering
    \includegraphics[width=1\linewidth]{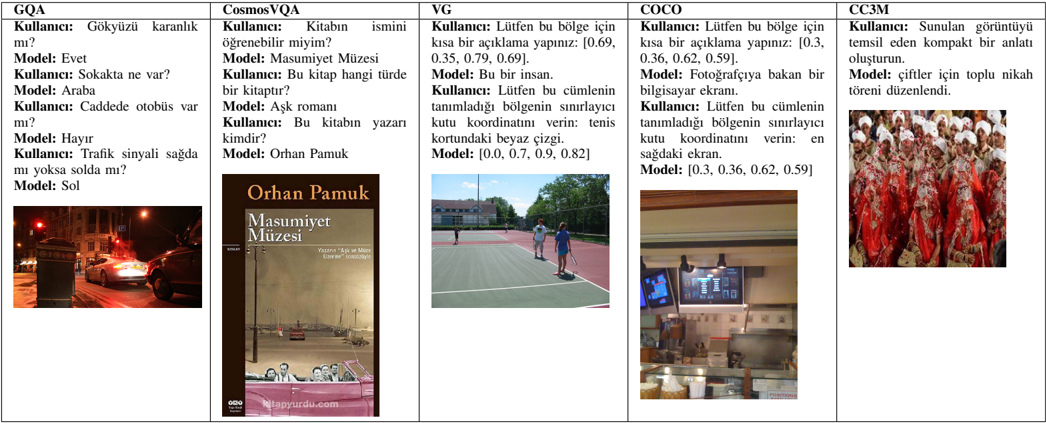}
    \caption{A figure of the model output which shows the misinterpretation of the table as an image, which should have been five separate images in a table, mixed with texts.}
    \label{fig:figure_errors_figure}
\end{figure*}
\appendix

\section{Error Analysis}
\label{sec:appendix_a}

\paragraph{Texts}
Models tend to show the Turkish characters (such as ğ and ş) as two separate characters. For instance, they interpret ğ as a breve symbol followed by a g, and ş as a cedilla symbol followed by an s. For example, "BEŞERİ" is interpreted as "BE¸ SERİ" and "DOĞAL" is interpreted as "DO˘ GAL". Another observation is that the uppercase i character (İ, not I as in English) in Turkish, is interpreted as Ì by some the models. For example, "GSYİH" is interpreted as "GSYÌH". Even though we can solve this problem with post-processing, it might be  challenging to predict how other models than the ones we evaluate will behave when encountered with such characters. We observe that some duplicated letters are not taken into account by the models. For example, "Laplace'ın \textit{Kuvvet} Fonksiyonları" is interpreted as "Laplace'ın \textit{Kuvet} Fonksiyonları" (missing a "v"). Also, models tend to ignore whitespaces and inline math equations as well. For example, "\textit{$m$ ve $n$}’nin reel sayı olması" is interpreted as "\textit{mven}’nin reel sayı olması".

\paragraph{Equations}
We observe incorrect variables and missing conditions for the partial expressions in the models. For example,
The expression
$$
\int_{a}^{b} f(t) \Delta t = \begin{cases} \sum_{k = \frac{a}{h}}^{\frac{b}{h}-1} f(kh)h, & \red{a < b \text{ ise,}} \\ 0, & \red{a = b \text{ ise,}} \\ -\sum_{k = \frac{\red{b}}{h}}^{\frac{a}{h}-1} f(kh)h, & \red{a > b \text{ ise}} \end{cases}
$$
is interpreted as
$$
\int_a^bf(t)\Delta t=\left\{\begin{array}{ c c }
\sum_{k=\frac{a}{h}}^{\frac{b}{h}-1}f(kh)h, &  \\
0, &  \\
-\sum_{k=\frac{\red{a}}{h}}^{\frac{a}{h}-1}f(kh)h, & 
\end{array}\right.
$$

\paragraph{Tables}
We encounter structural mistakes in the tables of the model outputs. These mistakes include incorrect split of the cells, unnecessary rows or columns, wrongly merged rows or columns. An example of the structural mistake is shown in Figure~\ref{fig:table_errors_figure}.

\paragraph{Figures}
Models interpret multiple figures as a one complete figure. An example of this figure combination error is shown in Figure~\ref{fig:figure_errors_figure}. Here the original PDF includes 5 images in a table. However, the model interpreted as a whole image.

\end{document}